\documentclass[letterpaper, 10 pt, conference]{ieeeconf}  %
\usepackage{amssymb,amsfonts}
\usepackage[linesnumbered,ruled,vlined]{algorithm2e}

\usepackage{graphicx}
\usepackage{flushend}
\usepackage{subfigure}
\graphicspath{{images/}}
\usepackage{graphics,multirow}
\usepackage{threeparttable}
\usepackage{color}
\usepackage{mathrsfs}
\usepackage{array}
\usepackage{booktabs}
\usepackage{xcolor}
\usepackage{tikz}

\usepackage{amssymb}
\usepackage{amsmath}

\usepackage{url}

\usetikzlibrary{arrows,shapes,chains}

\usepackage{algorithmic}

\newcommand{\qed}{\hfill\ensuremath{\square}}

\IEEEoverridecommandlockouts                              %

\title{\LARGE \bf
Grid-based Submap Joining: An Efficient Algorithm for Simultaneously Optimizing Global Occupancy Map and Local Submap Frames}

\author{Yingyu Wang\textsuperscript{1}, Liang Zhao\textsuperscript{2}, and Shoudong Huang\textsuperscript{1}  %
\thanks{$^{1}$Yingyu Wang and Shoudong Huang are with the Robotics Institute, Faculty of Engineering and Information Technology, University of Technology Sydney, Australia (e-mail: Yingyu.Wang-1@student.uts.edu.au; Shoudong.Huang@uts.edu.au).}
\thanks{$^{2}$Liang Zhao is with the School of Informatics, University of Edinburgh, Edinburgh, UK (e-mail: Liang.Zhao@uts.edu.au).}
}

\begin{document}

\maketitle
\thispagestyle{empty}
\pagestyle{empty}

\begin{abstract}
Optimizing robot poses and the map simultaneously has been shown to provide more accurate SLAM results. However, for non-feature based SLAM approaches, directly optimizing all the robot poses and the whole map will greatly increase the computational cost, making SLAM problems difficult to solve in large-scale environments. To solve the 2D non-feature based SLAM problem in large-scale environments more accurately and efficiently, we propose the grid-based submap joining method. Specifically, we first formulate the 2D grid-based submap joining problem as a non-linear least squares (NLLS) form to optimize the global occupancy map and local submap frames simultaneously. We then prove that in solving the NLLS problem using Gauss-Newton (GN) method, the increments of the poses in each iteration are independent of the occupancy values of the global occupancy map. Based on this property, we propose a pose-only GN algorithm equivalent to full GN method to solve the NLLS problem. The proposed submap joining algorithm is very efficient due to the independent property and the pose-only solution. Evaluations using simulations and publicly available practical 2D laser datasets confirm the outperformance of our proposed method compared to the state-of-the-art methods in terms of efficiency and accuracy, as well as the ability to solve the grid-based SLAM problem in very large-scale environments. The code is made available to benefit the robotics community\footnote{\url{https://github.com/WANGYINGYU/Occupancy-Joining}}.
\end{abstract}

\section{Introduction}

Occupancy grid map (OGM) is a widely used map representation in 2D environments because it categorizes the cells in the map into occupied, free, and unknown according to the presence or absence of obstacles in the corresponding environment \cite{moravec1985high, moravec1989sensor, elfes1989occupancy,hornung2013octomap,huang2024perception}, which is valuable for robot navigation and path planning.

Early works using OGM in 2D laser simultaneous localization and mapping (SLAM) techniques are particle filter based. Such techniques, including FastSLAM \cite{montemerlo2002fastslam} and GMapping \cite{grisetti2005improving, grisetti2007improved}, have the advantage that the occupancy of each cell could be integrated into the state representation, it is made possible as each particle includes both a robot trajectory and an associated OGM. However, due to particle filters maintaining the representation of the full system state in each particle, it is inevitable that particle filter-based methods will take up a lot of memory consumption and computation cost when the robot's trajectory becomes longer.

Optimization-based 2D laser SLAM methods perform better in terms of accuracy and resource consumption. One challenge in such approaches is to avoid cumulative errors as the robot trajectory grows. Hector SLAM \cite{kohlbrecher2011flexible} avoids the accumulation of errors by taking the scan-to-map matching method at the front-end of SLAM framework. Karto-SLAM \cite{konolige2010efficient} further introduces loop closure detection and implements global optimization using sparse pose adjustment to reduce the long-term error accumulation. Cartographer \cite{hess2016real} additionally introduces branch and bound strategy to speed up the loop closure detection. SLAM Toolbox \cite{macenski2021slam} integrated in ROS 2 is developed based on Karto-SLAM, which uses Ceres \cite{Agarwal_Ceres_Solver_2022} as the pose graph solver instead of sparse pose adjustment to provide faster and more flexible optimization settings. Moreover, SLAM Toolbox refactors scan matching method and introduces K-D tree search to speed up the computation.

Typically, as in Cartographer, optimization-based methods solve the SLAM problem in two steps. First, the robot poses are optimized using a pose graph, and then the optimized poses are assumed to be the correct poses and used to build up the OGM. However, in these two-step approaches, the uncertainties of the robot poses obtained in the first step are not taken into account when building the map, resulting in suboptimal solutions as pointed in \cite{Zhao-RSS-22,dellaert2006square,rosinol2021kimera,reijgwart2019voxgraph}.

Our recent work, Occupancy-SLAM \cite{Zhao-RSS-22}, demonstrates a substantial improvement in accuracy compared to two-step approaches by simultaneously optimizing the robot poses and the occupancy map. Specifically, it performs a sampling strategy on all laser beams to collect sample points as observations and then constructs error terms between the observations and the global occupancy map. In this approach, both the robot poses and the global occupancy map are considered as optimization variables and the proposed non-linear least squares (NLLS) problem is solved by a variant of the Gauss-Newton (GN) method.

Although Occupancy-SLAM has a great advantage in terms of accuracy, its computational complexity is related to the length of the robot trajectory and the map size because it optimizes all robot poses and the occupancy values of all the cells in the 2D grid based map together. It inevitably falls into the computational bottleneck and potential local minimum as the number of poses and size of the occupancy map grows. This limits the ability of Occupancy-SLAM to solve SLAM problems in large-scale environments.

Local submap joining is a commonly used scheme for SLAM in large-scale environments, because of its efficiency and less chance of being trapped in a local minimum as compared with full optimization based SLAM. Feature-based submap joining approaches \cite{zhao2013linear,wang2019submap} optimize local map coordinate frames and the feature map simultaneously and can achieve high-level of accuracy. However, existing OGM based submap joining approaches, such as Karto-SLAM \cite{konolige2010efficient}, SLAM Toolbox \cite{macenski2021slam} and Cartographer \cite{hess2016real}, only optimize poses but not the grid map, thus the accuracy is sacrificed to some extent. One challenge with OGM based submap joining methods is that optimizing poses and the global map simultaneously greatly affects computational efficiency due to the huge global occupancy map in large-scale environments.

\textbf{Contribution.} This paper considers the grid-based submap joining problem. We propose an efficient pose-only GN method which is equivalent to full GN to solve a NLLS problem where both the global map and the submap frames are optimized together. This is achieved by first proving an independent property of the full GN method for solving the NLLS. Specifically, the contributions of the paper are
\begin{enumerate}
	\item We formulate the grid-based submap joining problem as a NLLS problem where both local submap coordinate frames and the global occupancy map are considered as state variables to be optimized simultaneously.
	\item We prove that, when solving the formulated NLLS for both poses and the global map using GN method, the increment of poses in each iteration is independent of the occupancy values of the global occupancy map.

	\item Based on the independent property, we propose a pose-only GN method which is equivalent to full GN method for solving the proposed formulation. The optimal global occupancy map can be obtained using a closed-form formula after the optimal poses are obtained. Experiments demonstrate that the pose-only GN algorithm is very efficient and much faster than the full GN method.

	\item Experimental results using simulated and practical datasets confirm the superior performance of the proposed method in terms of efficiency and accuracy compared to existing state-of-the-art approaches.
\end{enumerate}

\section{Problem Formulation: \\Local Grid-based Submap Joining }\label{sec:problem}

\begin{figure}[t]
\centering
\vspace{0.52em}
\includegraphics[width=0.48\textwidth]{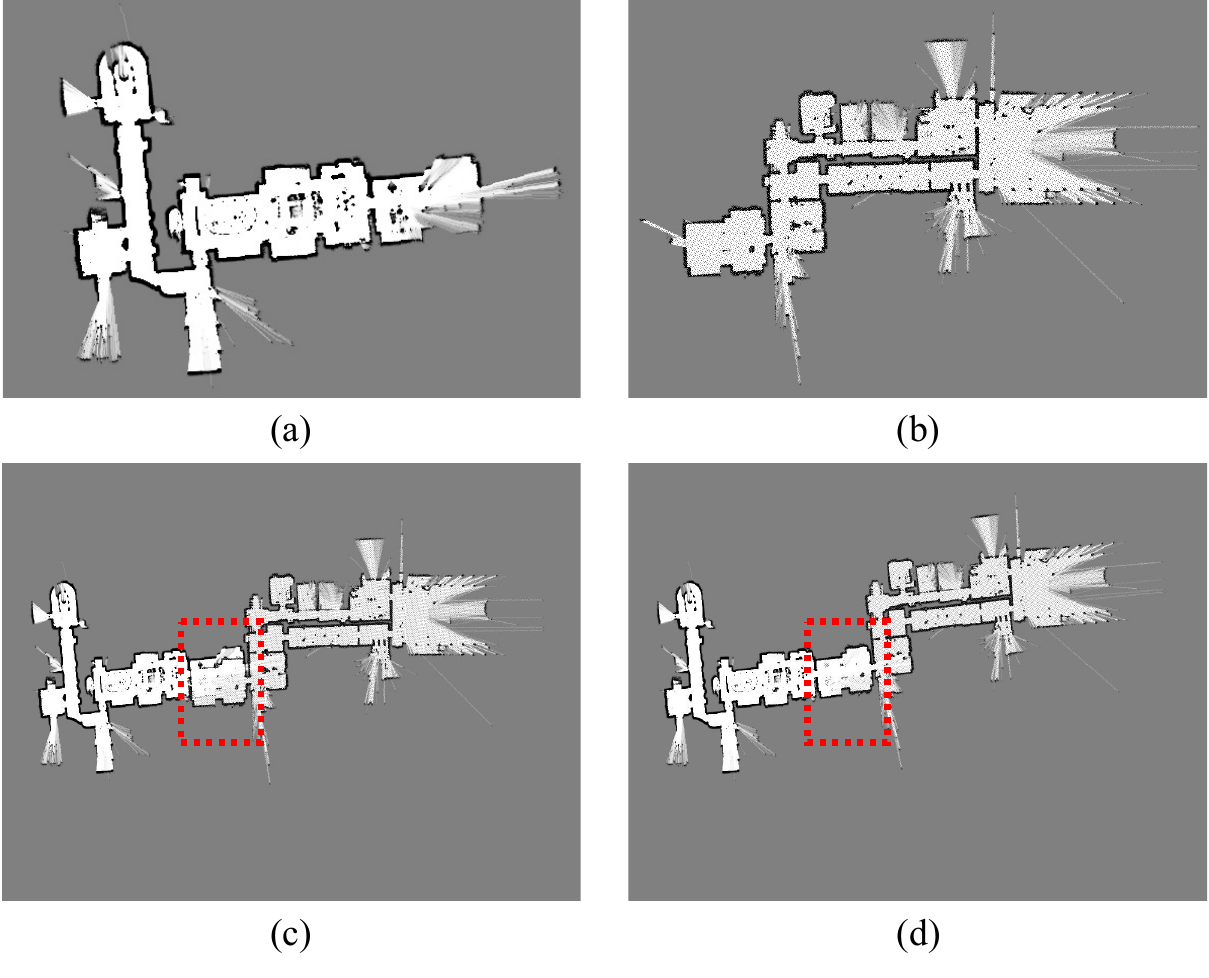}
\caption{\label{fig_demo} \textit{The description of the grid-based submap joining problem.} (a) and (b) show the grid-based submaps. (c) shows the global grid map generated by fusing the two submaps using the pose information only. (d) shows the consistent global grid map generated by our proposed grid-based submap joining method.}
\end{figure}

In this section, different from \cite{hess2016real,macenski2021slam}, we formulate the grid-based submap joining problem as a NLLS problem in which the variables are not the robot poses, but the global grid map and the local submap frames. A description of the local grid-based submap joining problem is shown in Fig. \ref{fig_demo}.

\subsection{Inputs and Outputs of Submap Joining Problem}
For the local submap joining problem, the input is a sequence of grid-based local submaps. First, let us denote $n+1$ submaps as $\boldsymbol{L} = \{\boldsymbol{L}_0, \cdots, \boldsymbol{L}_n\}$, where $\boldsymbol{L}_i$ represents the $i$-th local occupancy map built by any evidence grid mapping technique such as \cite{elfes1989occupancy,hess2016real,Zhao-RSS-22}. Thus, occupancy values of cells in these submaps are the logarithm of odds (the ratio between the probability of being occupied and the probability of being free). Different from \cite{hess2016real}, we do not maintain all the robot poses in each submap. The only poses we care about are the submap frames. Assume that the coordinate frames of these local submaps in the global map coordinate frame are $\left\{\boldsymbol{X}^r_0, \ldots, \boldsymbol{X}^r_n\right\} \in {SE}(2)^{n+1}$, where $\boldsymbol{X}^r_i = [\boldsymbol{t}_i^T, \theta_i]^T$ represents the $i$-th local submap coordinate frame. $\boldsymbol{t}_i$ is the $x$-$y$ position and $\theta_i$ is the orientation with the corresponding rotation matrix $\boldsymbol{R}_i = \left[ \begin{array}{rc} \cos(\theta_i) & \sin(\theta_i) \\ -\sin(\theta_i) & \cos(\theta_i) \end{array} \right ]$.

The task of our submap joining problem is to find the optimal poses of local submap coordinate frames and the optimal global grid map. We denote the global grid map as $ \{\boldsymbol{M}(\boldsymbol{m}_1), \cdots, \boldsymbol{M}(\boldsymbol{m}_{l_w \times l_h} ) \}$, where $\boldsymbol{m}_{j}~(1 \leq j \leq l_w \times l_h)$ represents the coordinate of a discrete cell in the global map and $\boldsymbol{M}(\boldsymbol{m}_{j})$ represents its corresponding occupancy value \cite{martin1996robot}. Therefore, the outputs of submap joining problem are the optimal solution of local submap coordinate frames and the optimal global map.

\subsection{Global to Local Coordinate Projection}
The grid cell $\boldsymbol{m}_{j}$ in the global map $\boldsymbol{M}$ can be projected to $i-$th local submap coordinate by pose $\boldsymbol{X}_i^r$, i.e.,
\begin{equation}
	\boldsymbol{p}_{im_{j}} = \boldsymbol{R}_i (\boldsymbol{m}_{j} \cdot s  - \boldsymbol{t}_i), \label{eq_p}
\end{equation}
where $\boldsymbol{p}_{im_{j}}$ is the projected position in the local submap coordinate, and $s$ is the resolution of cells in the global occupancy map $\boldsymbol{M}$ (the distance between two adjacent cells represents $s$ meters in the real world).

\subsection{The NLLS Formulation \label{sec_NLLS}}

The state vector of the proposed map joining problem is
\begin{equation}
	\boldsymbol{X} = [(\boldsymbol{X}^r)^T, \boldsymbol{M}^T]^T,
\end{equation}
where
\begin{equation}
\begin{aligned}
\boldsymbol{X}^r & =\left[\left(\boldsymbol{X}_1^r\right)^T, \cdots,\left(\boldsymbol{X}_n^r\right)^T\right]^T \\
\boldsymbol{M} & =\left[\boldsymbol{M}\left(\boldsymbol{m}_{1}\right), \cdots, \boldsymbol{M}\left(\boldsymbol{m}_{l_w\times l_h}\right)\right]^T.
\end{aligned}
\end{equation}
As with most submap joining problem formulations, we fix the first local map coordinate frame as the global coordinate. Therefore, $\boldsymbol{X}^r$ consists of $n$ local map coordinate frames and $\boldsymbol{M}$ includes $l_w \times l_h$ discrete cells of global occupancy map.

By the global-to-local projection relationship, all cells of global occupancy map $\boldsymbol{M}$ can be projected to corresponding submaps to compute the difference in occupancy values to formulate the NLLS problem, i.e., minimize
\begin{equation}
	f(\boldsymbol{X}) =  \sum_{i=0}^n\sum_{ j \in \{obs_i\}} \left\| \omega_{i{m}_j} \boldsymbol{M}(\boldsymbol{m}_{j}) - \boldsymbol{\boldsymbol{L}}_i(\boldsymbol{p}_{im_{j}}) \right\|^2 \label{eq_NLLS},
\end{equation}
where $\{obs_i\}$ is the set including the indices of cells in the global occupancy map $\boldsymbol{M}$ projected onto local submap $\boldsymbol{L}_i$. $\boldsymbol{M}(\boldsymbol{m}_{j})$ refers to the occupancy value of the cell $\boldsymbol{m}_j$ in the global occupancy map $\boldsymbol{M}$. $\boldsymbol{L}_i(\boldsymbol{p}_{im_{j}})$ means the occupancy value of the continuous coordinate $\boldsymbol{p}_{im_{j}}$ in the local submap $\boldsymbol{L}_i$. Here we use bilinear interpolation to obtain the occupancy value $\boldsymbol{L}_i(\boldsymbol{p}_{im_{j}})$ on the discrete grid-based submap $\boldsymbol{L}_i$, similar to \cite{kohlbrecher2011flexible,Zhao-RSS-22}.

In (\ref{eq_NLLS}), $\omega_{i{m}_j}$ is the weight to establish an accurate relationship between the global map and local submaps w.r.t. occupancy values of corresponding coordinates because both local submaps and the global map are represented by the Bayesian approach \cite{ProbabilisticRobotics}. It can be calculated by
\begin{equation}
    \omega_{i{m}_j} = \frac{\boldsymbol{N}^{L_i}(\boldsymbol{p}_{im_j})}{\boldsymbol{N}^M(\boldsymbol{m}_j)}.
\end{equation}
Here, $\boldsymbol{N}^{L_i}$ is the local hit map associated with submap $\boldsymbol{L}_i$. Each cell in $\boldsymbol{N}^{L_i}$ describes the observation count of the corresponding cell in $\boldsymbol{L}_i$, which is similar to \cite{Zhao-RSS-22}. $\boldsymbol{N}^{L_i}(\boldsymbol{p}_{im_j})$ approximates the observation count of the continuous coordinate $\boldsymbol{p}_{im_{j}}$ using bilinear interpolation based on the observation count of surrounding discrete cells. $\boldsymbol{N}^M$ denotes the global hit map associated with $\boldsymbol{M}$ which is built by projecting all the local hit maps $\{\boldsymbol{N}^{L_i}\}$ into the global map coordinate frame through $\{\boldsymbol{X}_i^r\}$. Thus, $\omega_{i{m}_j}$ is a function of $\{\boldsymbol{X}_i^r\}$.

\section{Our Approach: \\Efficient Pose-only GN Algorithm }\label{sec:method}
In this section, we prove the existence of a special independent property when solving our NLLS formulation (\ref{eq_NLLS}) using GN method. Based on this special property, we propose a pose-only GN algorithm equivalent to full GN method to solve our NLLS problem.

\subsection{Iterative Solver to the NLLS Formulation}
Our submap joining problem is to seek $\boldsymbol{X}$ to minimize $f(\boldsymbol{X})$ in (\ref{eq_NLLS}). Typical iterative algorithms, such as GN method, can be used to solve (\ref{eq_NLLS}) iteratively, i.e., by starting with an initial guess $\boldsymbol{X}(0)$ and updating with $\boldsymbol{X}(k+1) = \boldsymbol{X}(k) + \Delta(k)$. The update vector $
\boldsymbol{\Delta}(k)=\left[\boldsymbol{\Delta}^r(k)^T, \boldsymbol{\Delta}^M(k)\right]^T$ is the solution to
\begin{equation}
\boldsymbol{J}^T \boldsymbol{J} \boldsymbol{\Delta}(k)=-\boldsymbol{J}^T {F}(\boldsymbol{X}(k)) \label{eq_solving}
\end{equation}
where $F(\boldsymbol{X})^TF(\boldsymbol{X}) = f(\boldsymbol{X})$ and $\boldsymbol{J}$ is the Jacobian matrix $\partial F / \partial \mathbf{X}$ evaluated at $\mathbf{X}(k)$. However, for grid-based submap joining problem, the number of cells $l_w \times l_h$ in the global map $\boldsymbol{M}$ is very large which leads to a high dimension of the state vector, so common methods to solve (\ref{eq_solving}) are time-consuming.

\subsection{A Special Independent Property of Our Formulation \label{sec_iii_b}}

Our NLLS formulation (\ref{eq_NLLS}) can be shown to have a special property, when optimizing poses and the global occupancy map together using GN method, the optimization of poses is independent of the global occupancy map.

\textbf{Proposition\label{Proposition}}. \textit{For each step of GN iteration for minimizing the NLLS problem in (\ref{eq_NLLS}), the increment of poses $\boldsymbol{\Delta}^r(k)$ is independent of the occupancy values of the global occupancy map $\boldsymbol{M}$.}

\textbf{Proof\label{Proof}}. The Jacobian matrix $\boldsymbol{J}$ consists of two parts, i.e. the Jacobian of $F(\boldsymbol{X})$ w.r.t. the poses $\boldsymbol{J}_r$, and the Jacobian of $F(\boldsymbol{X})$ w.r.t. the global occupancy map $\boldsymbol{J}_M$, i.e. $\boldsymbol{J} = [\boldsymbol{J}_r	\quad \boldsymbol{J}_M]$. All non-zero elements of $\boldsymbol{J}_M$ are $\omega_{i{m}_j}$ which only depend on poses (hit maps $\boldsymbol{N}^M$, $\boldsymbol{N}^{L_i}$ are independent of occupancy values), and thus $\boldsymbol{J}_M$ is independent of $\boldsymbol{M}$. In addition, it can be deducted from (\ref{eq_p}) and (\ref{eq_NLLS}) that $\boldsymbol{J}_r$ is also independent of $\boldsymbol{M}$.

Let us rewrite (\ref{eq_solving}) as
\begin{equation}
\left[\begin{array}{cc}
\boldsymbol{U} & \boldsymbol{W} \\
\boldsymbol{W}^T & \boldsymbol{V}
\end{array}\right]\left[\begin{array}{l}
\boldsymbol{\Delta}^r \\
\boldsymbol{\Delta}^M
\end{array}\right]=\left[\begin{array}{l}
\boldsymbol{b}^r \\
\boldsymbol{b}^M
\end{array}\right] \label{eq_schur}
\end{equation}
where $\boldsymbol{U} = \boldsymbol{J}_r^T \boldsymbol{J}_r$, $\boldsymbol{V} = \boldsymbol{J}_M^T \boldsymbol{J}_M$, $\boldsymbol{W} = \boldsymbol{J}_r^T  \boldsymbol{J}_M$, $\boldsymbol{b}^r = -\boldsymbol{J}_r^T F(\boldsymbol{X})$, and $\boldsymbol{b}^M = - \boldsymbol{J}_M^T F(\boldsymbol{X})$. According to the theory of the Schur complement \cite{zhang2006schur}, the solution of (\ref{eq_schur}) can be calculated by
\begin{equation}
\left(\boldsymbol{U}-\boldsymbol{W} \boldsymbol{V}^{-1} \boldsymbol{W}^T\right) \boldsymbol{\Delta}^r=\boldsymbol{b}^r-\boldsymbol{W} \boldsymbol{V}^{-1} \boldsymbol{b}^M \label{eq_delta_r}
\end{equation}
and
\begin{equation}
\boldsymbol{V} \boldsymbol{\Delta}^M=\boldsymbol{b}^M-\boldsymbol{W}^T \boldsymbol{\Delta}^r .
\end{equation}
If we rewrite (\ref{eq_delta_r}) as
\begin{equation}
\begin{aligned}
& \left(\boldsymbol{J}_r^T \boldsymbol{J}_r-\boldsymbol{J}_r^T  \boldsymbol{J}_M\left(\boldsymbol{J}_M^T  \boldsymbol{J}_M\right)^{-1} \boldsymbol{J}_M^T  \boldsymbol{J}_r\right) \boldsymbol{\Delta}^r \\
= & -\boldsymbol{J}_r^T F\left(\boldsymbol{X}\right) +\boldsymbol{J}_r^T  \boldsymbol{J}_M\left(\boldsymbol{J}_M^T  \boldsymbol{J}_M\right)^{-1} \boldsymbol{J}_M^T F\left(\boldsymbol{X}\right), \label{eq_proof_1}
\end{aligned}
\end{equation}
it can be seen that the left-hand side of (\ref{eq_proof_1}) is not related to $\boldsymbol{M}$ because both $\boldsymbol{J}_M$ and $\boldsymbol{J}_r$ are independent of $\boldsymbol{M}$.

Furthermore, we can rewrite (\ref{eq_NLLS}) as
\begin{equation}
    f(\boldsymbol{X})=\|F(\boldsymbol{X})\|^2 =\|\boldsymbol{J}_M\boldsymbol{M} - H(\boldsymbol{X}^r))\|^2, \label{eq_rewrite}
\end{equation}
where $H(\boldsymbol{X}^r) = [\cdots,\boldsymbol{\boldsymbol{L}}_i(\boldsymbol{p}_{im_{j}}),\cdots]$ is not related to $\boldsymbol{M}$. Using $F(\boldsymbol{X})$ in (\ref{eq_rewrite}), the right-hand side of (\ref{eq_proof_1}) becomes
\begin{equation}
    \begin{aligned}
      &  -\boldsymbol{J}_r^T(\boldsymbol{J}_M\boldsymbol{M} - \boldsymbol{J}_M (\boldsymbol{J}_M^T \boldsymbol{J}_M)^{-1}\boldsymbol{J}_M^T\boldsymbol{J}_M\boldsymbol{M} )\\
      & -\boldsymbol{J}_r^T(\boldsymbol{J}_M(\boldsymbol{J}_M^T\boldsymbol{J}_M)^{-1}\boldsymbol{J}_M^T H(\boldsymbol{X}^r) - H(\boldsymbol{X}^r)).
    \end{aligned} \label{eq_proof_2}
\end{equation}
It can be deduced that the first item of (\ref{eq_proof_2}) is equal to 0, and the second item is not related to $\boldsymbol{M}$ because $\boldsymbol{J}_M$, $\boldsymbol{J}_r$ and $H(\boldsymbol{X}^r)$ are all independent of $\boldsymbol{M}$. Thus, the right-hand side of (\ref{eq_proof_1}) is not related to $\boldsymbol{M}$. Therefore, the increment of poses $\boldsymbol{\Delta}^r$ is independent of $\boldsymbol{M}$ and it can be calculated by
\begin{equation}
\begin{aligned}
  &      \left(\boldsymbol{J}_r^T\boldsymbol{J}_r-\boldsymbol{J}_r^T\boldsymbol{J}_M (\boldsymbol{J}_M^T\boldsymbol{J}_M)^{-1} \boldsymbol{J}_M^T\boldsymbol{J}_r\right) \boldsymbol{\Delta}^r \\
 &   = \boldsymbol{J}_r^T H(\boldsymbol{X}^r) - \boldsymbol{J}_r^T\boldsymbol{J}_M(\boldsymbol{J}_M^T\boldsymbol{J}_M)^{-1}\boldsymbol{J}_M^TH(\boldsymbol{X}^r). \label{eq_solve_delta_r}
    \end{aligned}
\end{equation}

For the point feature based SLAM problem with known data association and under the assumption that the observation noises covariance matrices are isotropic, a similar property was proved in \cite{Zhao-RSS-21}. In this paper, we prove that this independent property also holds in the proposed grid-based submap joining problem without the assumption of data association.  \qed

\begin{algorithm}[tp]
 \caption{Our Pose-only GN Algorithm}\label{alg_1}
 \begin{algorithmic}[1]
   \REQUIRE
     Submaps $\{\boldsymbol{L_i}\}$, local hit maps $\{\boldsymbol{N}^{L_i}\}$, and initial poses $\boldsymbol{X}^r(0)$
   \ENSURE
    Optimized poses $\hat{\boldsymbol{X}}^r$ and the optimized global occupancy map $\hat{\boldsymbol{M}}$
   \STATE Initialize global occupancy map $\boldsymbol{M}(0)$, global hit map $\boldsymbol{N}^M(0)$ using $\boldsymbol{X}^r(0)$, $\{\boldsymbol{L_i}\}$ and $\{\boldsymbol{N}^{L_i}\}$

   \FOR {$k=0$; $k <= \tau_k \; \& \; \| \boldsymbol{\Delta}^r(k) \|^2 >= \tau^r_{\Delta}$; $k++$}
     \STATE Evaluate $H(\boldsymbol{X}^r)$ and Jacobian $\boldsymbol{J}_r$, $\boldsymbol{J}_M$ at $\boldsymbol{X}^r(k)$
     \STATE Solve linear system (\ref{eq_solve_delta_r}) to get $\boldsymbol{\Delta}^r(k)$
     \STATE Update poses $\boldsymbol{X}^r(k+1)=\boldsymbol{X}^r(k) + \boldsymbol{\Delta}^r(k)$
     \STATE Recalculate the global hit map $\boldsymbol{N}^M(k+1)$ using $\boldsymbol{X}^r(k+1)$ and $\{\boldsymbol{N}^{L_i}\}$
   \ENDFOR
    \STATE $\hat{\boldsymbol{X}}^r = \boldsymbol{X}^r(k+1)$
   \STATE  Solve closed-form formula (\ref{eq_solve_map}) to get the optimized global occupancy map $\hat{\boldsymbol{M}}$
 \end{algorithmic}
\end{algorithm}
\setlength{\textfloatsep}{3pt}
\subsection{Equivalent Pose-only GN Iteration Algorithm}

According to the Proposition proved in Section \ref{sec_iii_b}, we propose a pose-only GN algorithm equivalent to full GN method to solve our NLLS problem. The equivalent pose-only GN algorithm is given in Algorithm \ref{alg_1}.
In Algorithm \ref{alg_1}, we iteratively solve (\ref{eq_solve_delta_r}) using Jacobian $\boldsymbol{J}_r$ and $\boldsymbol{J}_M$ to obtain the increments of the poses $\boldsymbol{\Delta}^r$ and update poses, and then recalculate the global hit map $\boldsymbol{N}^M$ based on the updated poses since $\boldsymbol{J}_M$ depends on the global hit map. After the optimal solution of poses $\boldsymbol{\hat{X}}^r$ is obtained, our NLLS problem (\ref{eq_NLLS}) becomes a linear least squares problem which minimize
\begin{equation}
g(\boldsymbol{M}) = (\boldsymbol{J}_M \boldsymbol{M} - H(\hat{\boldsymbol{X}}^r))^T (\boldsymbol{J}_M \boldsymbol{M} - H(\hat{\boldsymbol{X}}^r)).
\end{equation}
Therefore, the optimal solution of the global occupancy map $\hat{\boldsymbol{M}}$ can be calculated by the closed-form formula
\begin{equation}
     \hat{\boldsymbol{M}} = \boldsymbol{V}^{-1}\boldsymbol{J}_MH(\hat{\boldsymbol{X}}^r). \label{eq_solve_map}
\end{equation}

For Algorithm \ref{alg_1}, the dimension of the sparse linear system to be solved at each iteration only depends on the number of poses, whereas in the full GN algorithm, it depends on the number of poses and the number of cells in the global map. For the grid-based submap joining problem in which the number of poses is much less than that of cells of the global map, our pose-only GN algorithm can easily and quickly obtain the solution of $\boldsymbol{\Delta}^r$ for each iteration. After optimal poses $\boldsymbol{\hat{X}}^r$ are obtained, the optimized global map $\hat{\boldsymbol{M}}$ can be easily obtained by the closed-form formula (\ref{eq_solve_map}).

It should be noted that, although our pose-only GN does not need to update the map during iterations, the constraint information associated with the map is still implicitly taken into account when optimizing the poses, thus our method is quite different from pose-graph optimization.

\section{Experimental Results}\label{sec:results}
In this section, we demonstrate the accuracy and efficiency of our algorithm. In the experiments, we compare our results with Cartographer \cite{hess2016real}, SLAM Toolbox \cite{macenski2021slam} and Occupancy-SLAM \cite{Zhao-RSS-22}, which are the current state-of-the-art algorithms and perform significantly better than other methods such as Hector-SLAM \cite{kohlbrecher2011flexible}, Karto-SLAM \cite{konolige2010efficient}, etc.

First, due to the lack of ground truth in the practical datasets, we evaluate our algorithm qualitatively and quantitatively using datasets generated from two simulated environments. Second, we validate and evaluate the performance of our method through six publicly available practical datasets, including three datasets in large-scale environments. Parameters of all used datasets are listed in Table \ref{tab_dataset}. All OGMs are generated by poses from different approaches and corresponding observations using the same evidence grid mapping technique. In addition, we use Occupancy-SLAM to build our local submaps for all the experiments, in which the map resolution is the default setting in \cite{Zhao-RSS-22} for simulation experiments and low-resolution for practical experiments.

 \begin{table}[th]
		\centering
		\caption{Parameters of Datasets. \label{tab_dataset}}
		\setlength{\tabcolsep}{0.9mm}{
		\begin{tabular}{lcccc}\toprule
		Dataset	& No. Scans & Duration (s)  & Map Size (m) &  Odometry\\ \hline
		Simulation 1 & 3640  &117 & $50 \times 50$ & yes \\
		Simulation 2  & 2680  & 83 & $50 \times 50$ & yes  \\
		Car Park \cite{zhao20202d}  & 1642 & 164 & $50 \times 40$ & yes \\
		C5 \cite{Zhao-RSS-22} & 3870  &136 & $50 \times 40$ & yes\\
		Museum b0 1G \cite{hess2016real} & 5522 &152 & $85 \times 95$ &no  \\
		Museum b2 \cite{hess2016real} & 51833 &1390 &  $250 \times 200$ &no\\
        Museum b0 EG \cite{hess2016real} &22650 & 615 &  $225 \times 150$ & no \\
        C3 &24402 &610 & $150 \times 125$  & no\\
		\hline
		\end{tabular}
		}
\vspace{1.0em}
\end{table}

\subsection{Simulation Experiments}

We use varying levels of nonlinearity, non-convex obstacles, and long corridors to design two different simulation experiments. For all simulated datasets, each scan consists of 1081 laser beams with angles ranging from -135 degrees to 135 degrees which simulates a Hokuyo UTM-30LX laser scanner. To simulate real-world data acquisition, we add random Gaussian noises with zero-mean and standard deviation of 0.02 m to each beam of the scan data generated from the ground truth. Similarly, we add zero-mean Gaussian noises to the odometry inputs generated from the ground truth poses (standard deviation of 0.04 m for x-y and 0.003 rad for orientation). For each simulation environment, we generate 5 datasets with different sets of random noises.

\begin{figure}[t]
\centering \subfigure[Simulation 1] {\label{fig_trajectory_1}
\includegraphics[width=0.205\textwidth]{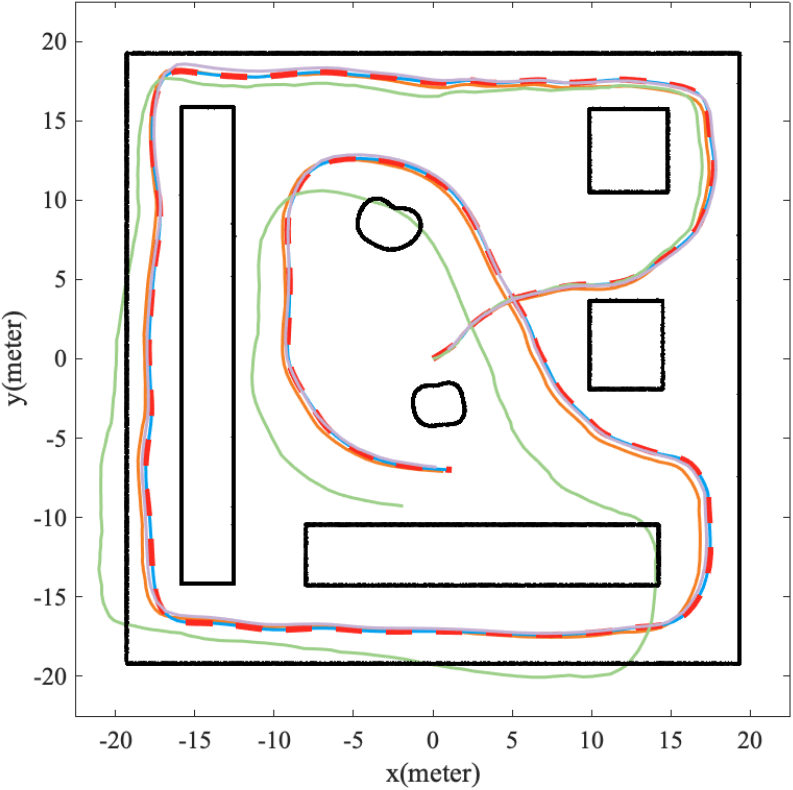}}
\centering \subfigure[Simulation 2] {\label{fig_trajectory_2}
\includegraphics[width=0.255\textwidth]{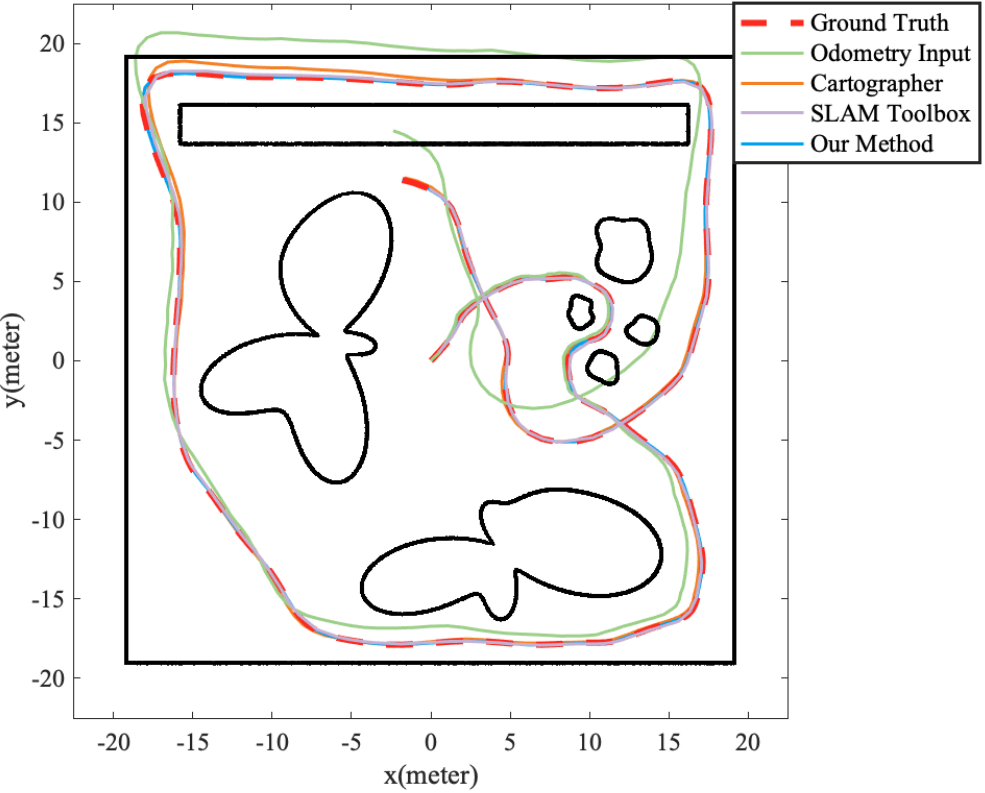}}
\caption{\label{fig_trajectory_compare} \textit{Simulation environments and robot trajectory results.} (a) and (b) show the simulation environments (the black lines indicate the obstacles in the scene) and the trajectories of ground truth, odometry inputs, Cartographer \cite{hess2016real}, SLAM Toolbox \cite{macenski2021slam}, and our method for one dataset in each of the two simulation experiments. Results from Occupancy-SLAM \cite{Zhao-RSS-22} are visually identical to our method and not drawn in the figure.}
\vspace{1.0em}
\end{figure}

The robot trajectory estimation results of our method, SLAM Toolbox and Cartographer of one dataset in each simulation are compared with the ground truth and odometry in Fig. \ref{fig_trajectory_compare}. Since no difference can be visualized between our trajectory and the trajectory from Occupancy-SLAM, the trajectory from Occupancy-SLAM is not drawn. It is obvious that our trajectories are closer to the ground truth trajectories.

\begin{figure*}[th]
\centering
\includegraphics[width=0.95\textwidth]{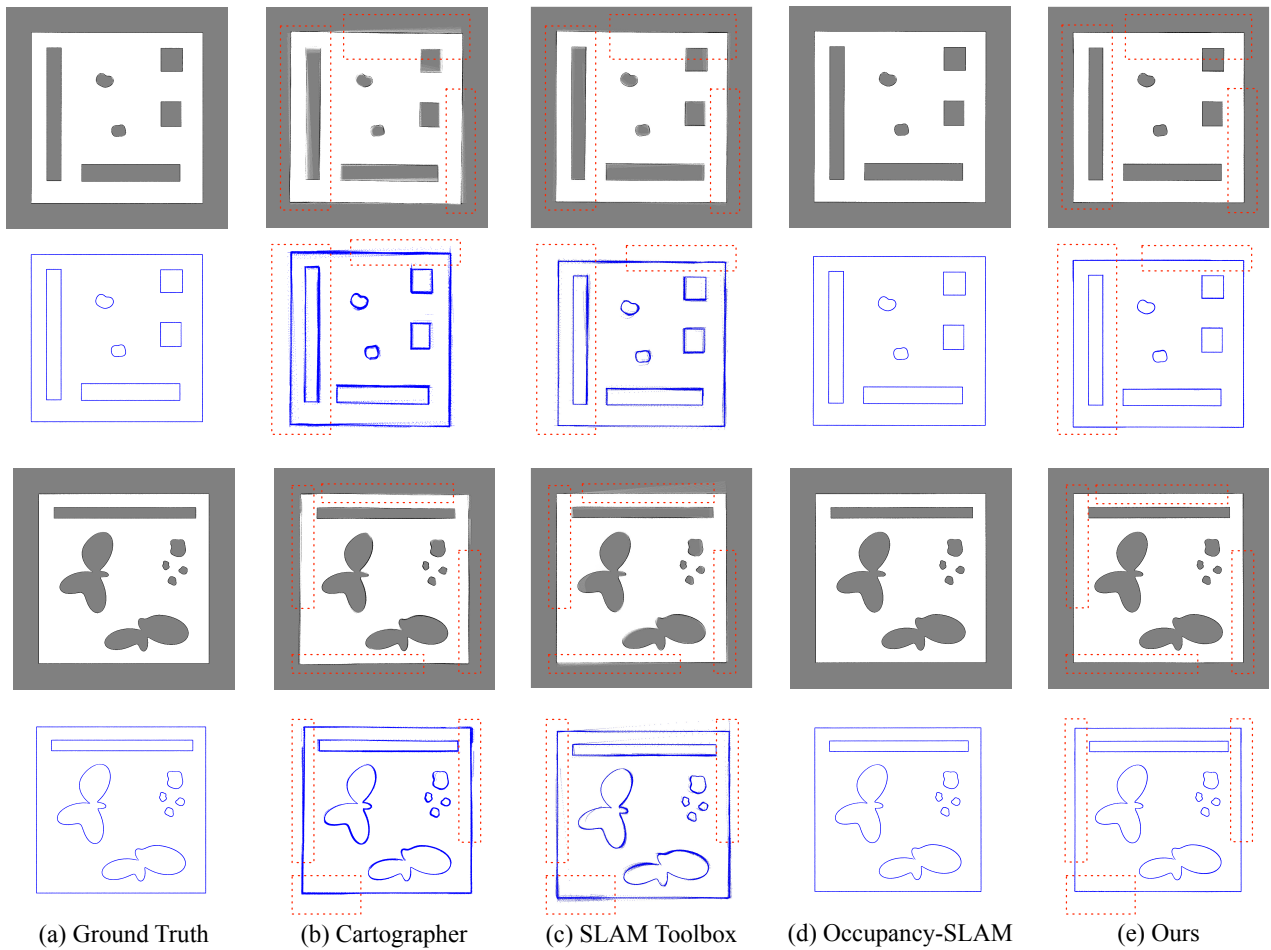}
\caption{\label{fig_simu} \textit{The OGMs and point cloud maps generated by poses from ground truth, Cartographer, SLAM Toolbox, Occupancy-SLAM and our approach in one dataset for Simulation 1 (first two rows) and Simulation 2 (last two rows).}}
\end{figure*}

The quantitative comparison of errors in the estimated poses from Cartographer, SLAM Toolbox, Occupancy-SLAM and our method is given in Table \ref{tab_comparison}, in which we use mean absolute error (MAE) and root mean squared error (RMSE) to evaluate the translation errors (in meters) and rotation errors (in radians) for the 5 runs in each simulation. It should be noted that for SLAM Toolbox, since only the poses of submap frames are involved in the pose graph optimization, here we only use those poses of the submap frames to evaluate the pose errors. Compared to Cartographer and SLAM Toolbox, our approach performs better on all metrics, with errors about 3 to 12 times smaller. The significant reduction in error validates the effectiveness of our proposed method. Occupancy-SLAM performs best in terms of accuracy due to the use of the full least squares strategy, which is the best one can achieve but is costly.

\begin{table}[h]
		\centering
		\caption{Comparison of Robot Pose Errors in Simulations.}
		\label{tab_comparison}
		\setlength{\tabcolsep}{0.6mm}{
		\begin{tabular}{lccccc}\toprule
			& Odom & Carto  & Toolbox &Occupancy-SLAM* & Ours  \\ \hline
		Simulation 1& & & &\\
		\quad MAE (Trans/m) & 0.7827 & 0.2534 & 0.1465&\textcolor{red} {0.0238} &\textcolor{blue}{0.0453} \\
		\quad MAE (Rot/rad) & 0.0491 & 0.0139 &0.0134 & \textcolor{red} {0.0008} & \textcolor{blue}{0.0012} \\
		\quad RMSE (Trans/m) & 0.9840 & 0.2992 & 0.1788& \textcolor{red} {0.0263} &\textcolor{blue}{0.0609} \\
		\quad RMSE (Rot/rad) & 0.0551 & 0.0156 &0.0152 & \textcolor{red} {0.0009} & \textcolor{blue}{0.0015} \\\hline
		Simulation 2& & & &\\
		\quad MAE (Trans/m) & 0.7535 & 0.1426 &0.0493 & \textcolor{red} {0.0070} & \textcolor{blue}{0.0151} \\
		\quad MAE (Rot/rad) & 0.0518 & 0.0068 &0.0085 & \textcolor{red} {0.0006} & \textcolor{blue}{0.0011} \\
		\quad RMSE (Trans/m) & 0.9687 & 0.1878 &0.0626 & \textcolor{red} {0.0106} &\textcolor{blue}{0.0205}\\
		\quad RMSE (Rot/rad) & 0.0593 & 0.0091 & 0.0127& \textcolor{red} {0.0009} & \textcolor{blue}{0.0018} \\\hline
		\end{tabular}
	\begin{tablenotes}
     \item \textcolor{red}{{Red}} and the \textcolor{blue}{{blue}} indicate the best and second best results respectively.
     \item * Occupancy-SLAM uses the full least squares strategy including all the robot poses, which can theoretically achieve the highest accuracy. However, it is very costly and cannot handle large-scale environments. For Simulation 1 and Simulation 2, Occupancy-SLAM takes about 20 times longer than our submap joining approach. The time consumptions on some practical datasets are given in Table. \ref{table_time_compare}.
   \end{tablenotes}
   }
   \vspace{2em}
\end{table}

Fig. \ref{fig_simu} shows the OGMs and point cloud maps generated by poses from ground truth, Cartographer, SLAM Toolbox, Occupancy-SLAM and our approach, where the first two rows are the results of a dataset in Simulation 1, and the last two rows are the results of a dataset in Simulation 2. The areas highlighted by red dots show that the results of our method are better than the results of Cartographer and SLAM Toolbox. It is clear that the object boundaries of both OGMs and point cloud maps obtained by our method are much clearer than Cartographer and SLAM Toolbox. Also, it can be seen that the maps of our method are very close to Occupancy-SLAM. This suggests that our submaps joining method, as an approximation scheme of full optimization based SLAM, can obtain similar results as the full optimization based approach Occupancy-SLAM.

For the quantitative comparison of occupancy maps, we use the results from all the ten datasets in Simulation 1 and Simulation 2 to compare the accuracy of the OGMs. We use AUC (Area under the ROC curve) \cite{bradley1997use} and precision to evaluate our method, Cartographer, SLAM Toolbox and Occupancy-SLAM, where ground truth labels are given by OGMs generated using ground truth poses. When evaluating using AUC, we remove all the unknown cells because AUC is a metric for binary classification as in \cite{bradley1997use}. The results are given in Table \ref{tab_auc}. Compared with Cartographer and SLAM Toolbox, it can be seen that our method achieves better performance in both metrics, which proves our approach can obtain more accurate maps.

\begin{figure}[t]
\centering
\includegraphics[width=\linewidth]{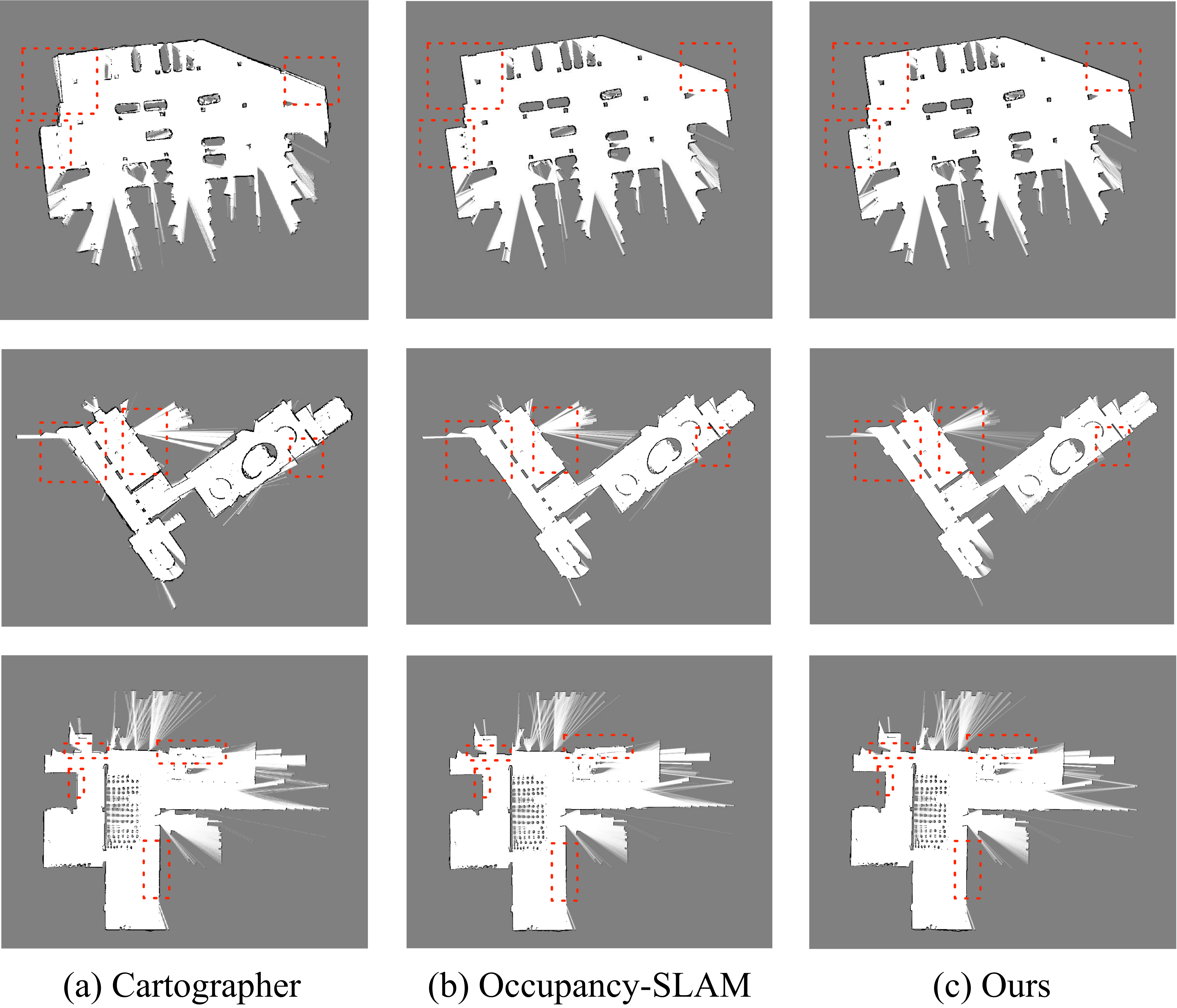}
\caption{\label{fig_result_compare_OGM} \textit{OGMs from Cartographer, Occupancy-SLAM and our method.} The first to third rows are \textit{Car Park}, \textit{Deutsches Museum b0 1G} and \textit{C5} dataset, respectively.}
\vspace{1.0em}
\end{figure}

\begin{figure}[t]
\centering
\includegraphics[width=\linewidth]{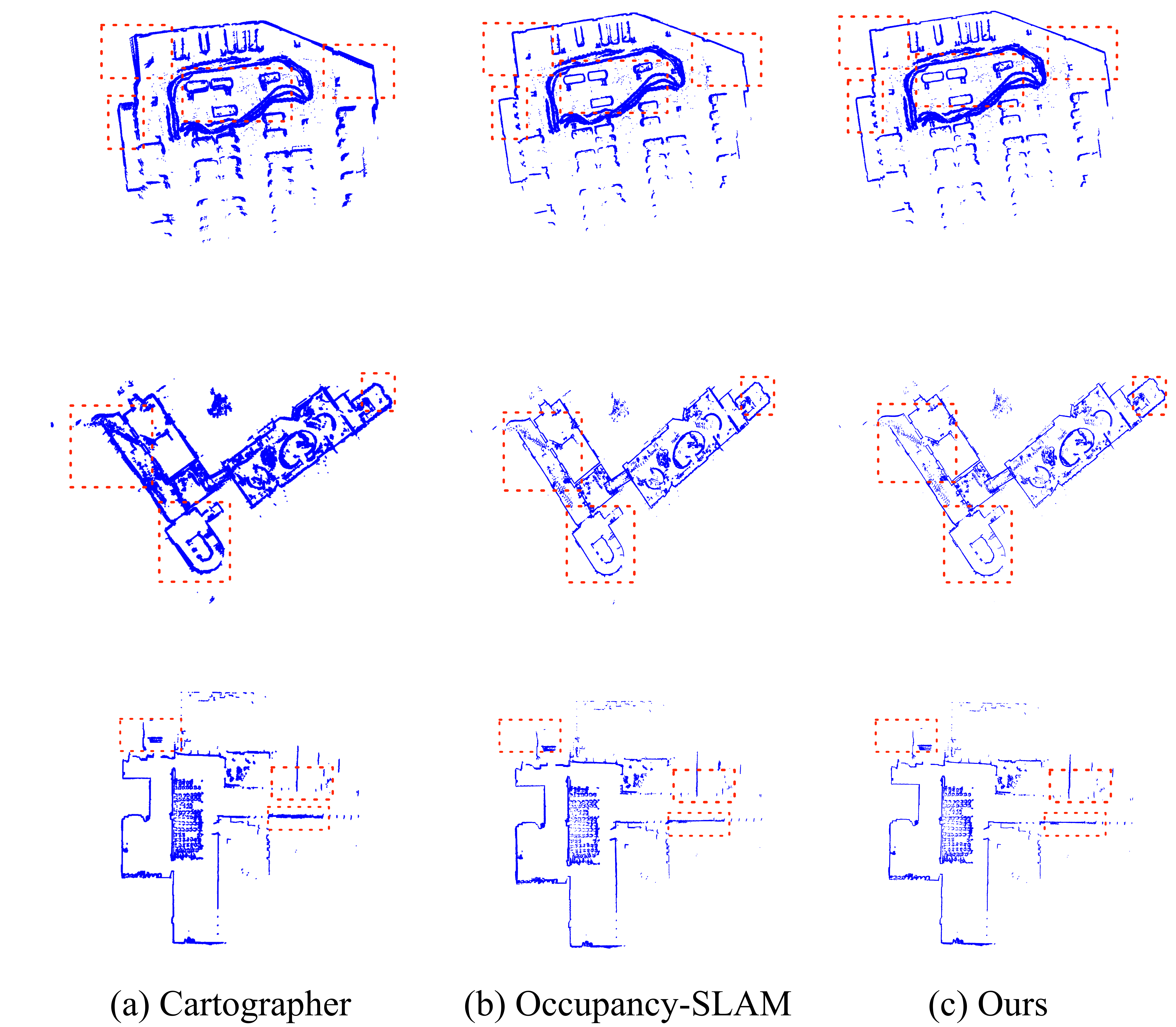}
\caption{\label{fig_result_compare_Scan} \textit{Point cloud maps from Cartographer, Occupancy-SLAM and our method.} The point cloud maps are generated by projecting the scan endpoints using poses.}
\vspace{1.0em}
\end{figure}

\begin{table}[thp]
		\centering
		\caption{Accuracy of the Occupancy Grid Maps.}
		\label{tab_auc}
		\setlength{\tabcolsep}{3.8 mm}{
		\begin{tabular}{llccccc}\toprule
		& & AUC & Precision   \\ \hline
		\multirow{4}{*}{Simulation 1} & Cartographer & 0.9088 &0.9565 \\ & SLAM Toolbox & 0.9550 &0.9749 \\ & Occupancy-SLAM & \textcolor{red}{{0.9997}} & \textcolor{red}{{0.9973}} \\ & Ours & \textcolor{blue}{{0.9751}} & \textcolor{blue}{{0.9867}} \\\hline

		\multirow{4}{*}{Simulation 2} & Cartographer & 0.9270 & 0.9660 \\ & SLAM Toolbox & 0.9726 &0.9871 \\& Occupancy-SLAM &\textcolor{red}{{0.9966}} &\textcolor{red}{{0.9991}}\\ & Ours & \textcolor{blue}{{0.9949}} & \textcolor{blue}{{0.9979}} \\ \hline
						\end{tabular}
		}
  \vspace{1.0em}
\end{table}

\subsection{Comparisons Using Practical Datasets}

First, we use three practical datasets used in \cite{Zhao-RSS-22}, namely \textit{Deutsches Museum b0 1G} \cite{hess2016real}, \textit{Car Park} \cite{zhao20202d} and \textit{C5} \cite{Zhao-RSS-22}, to compare our method with Cartographer and Occupancy-SLAM. The results are shown in Fig. \ref{fig_result_compare_OGM} and Fig. \ref{fig_result_compare_Scan}. It is clear that our results are very close to those of Occupancy-SLAM. Compared with Cartographer's results, both the OGMs and point cloud maps of our results have significantly clearer object boundaries. These results show the accuracy of our submap joining method using practical datasets. Since SLAM Toolbox does not support the \textit{MultiEchoLaserScan} sensor message format of Cartographer's companion
\textit{Deutsches Museum} dataset \cite{hess2016real} and performs poorly on \textit{Car Park} and \textit{C5} datasets because no loop closures are detected to perform pose graph optimization, we do not present the results of the SLAM Toolbox for this and subsequent experiments.

\begin{figure*}[h]
\centering \subfigure[b2 Cartographer] { \label{fig_large_carto_b2}
\includegraphics[width=0.308\textwidth]{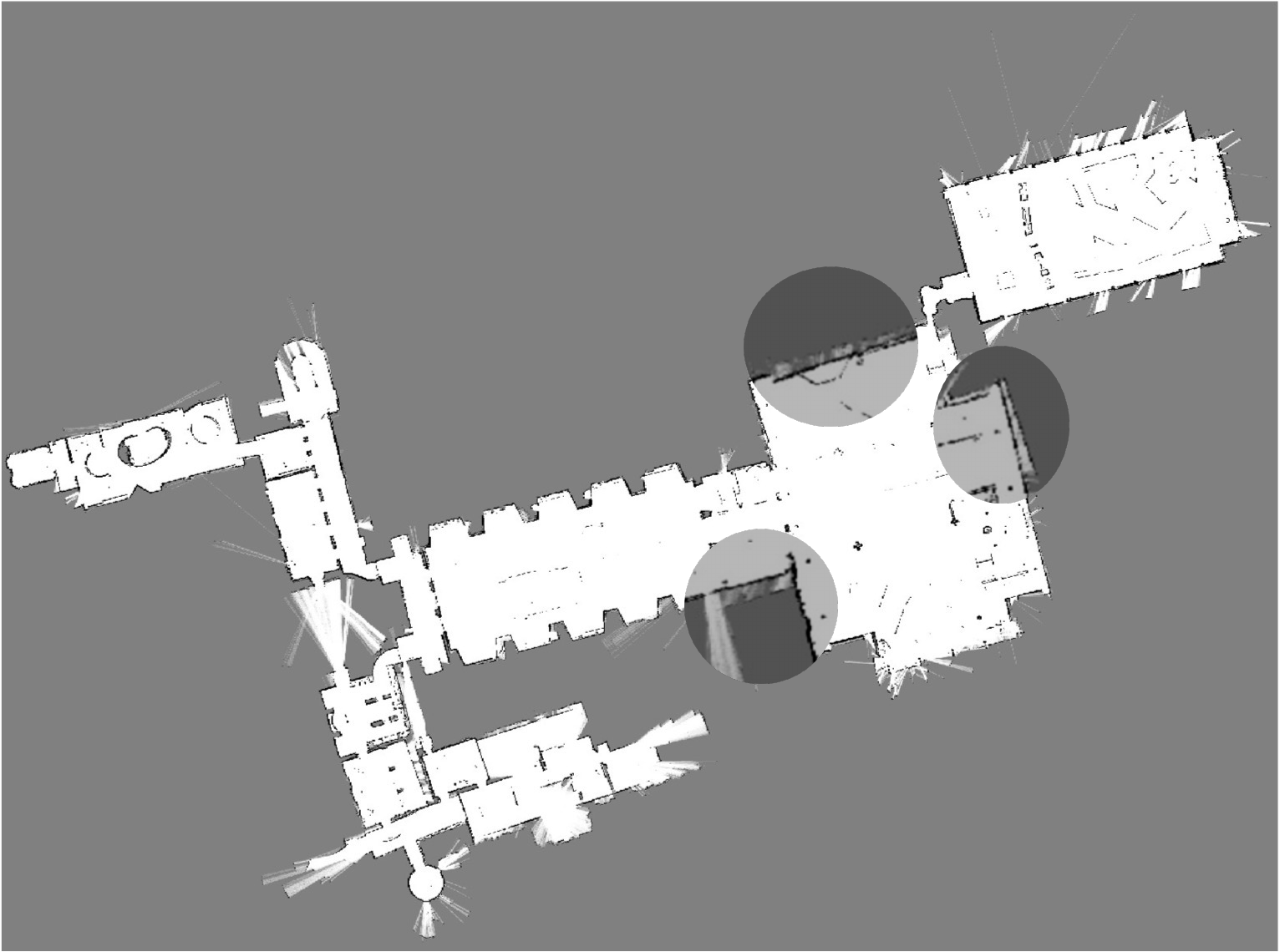}}
\vspace*{-0.6em}
\centering \subfigure[b0 EG Cartographer]{\label{fig_large_carto_b0}
\includegraphics[width=0.385\textwidth]{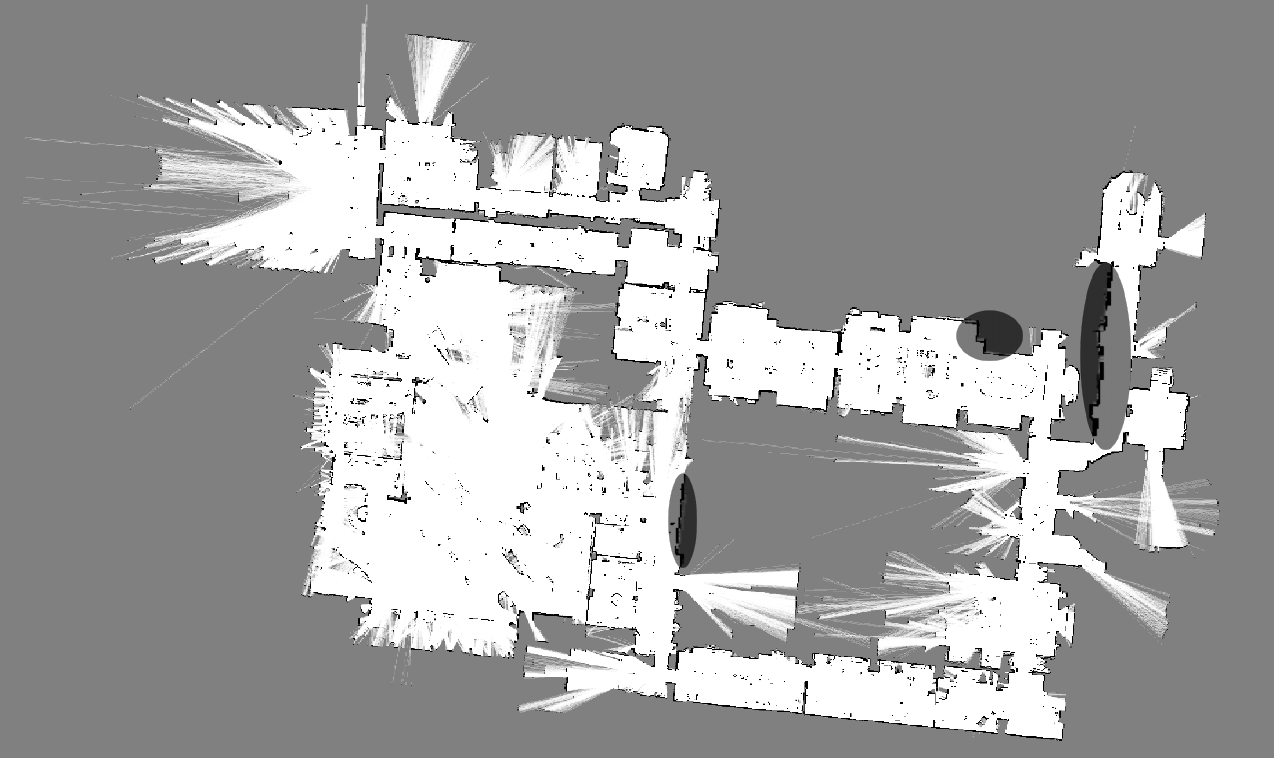}}
\centering \subfigure[C3 Cartographer] {\label{fig_large_C3_Carto}
\includegraphics[width=0.265\textwidth]{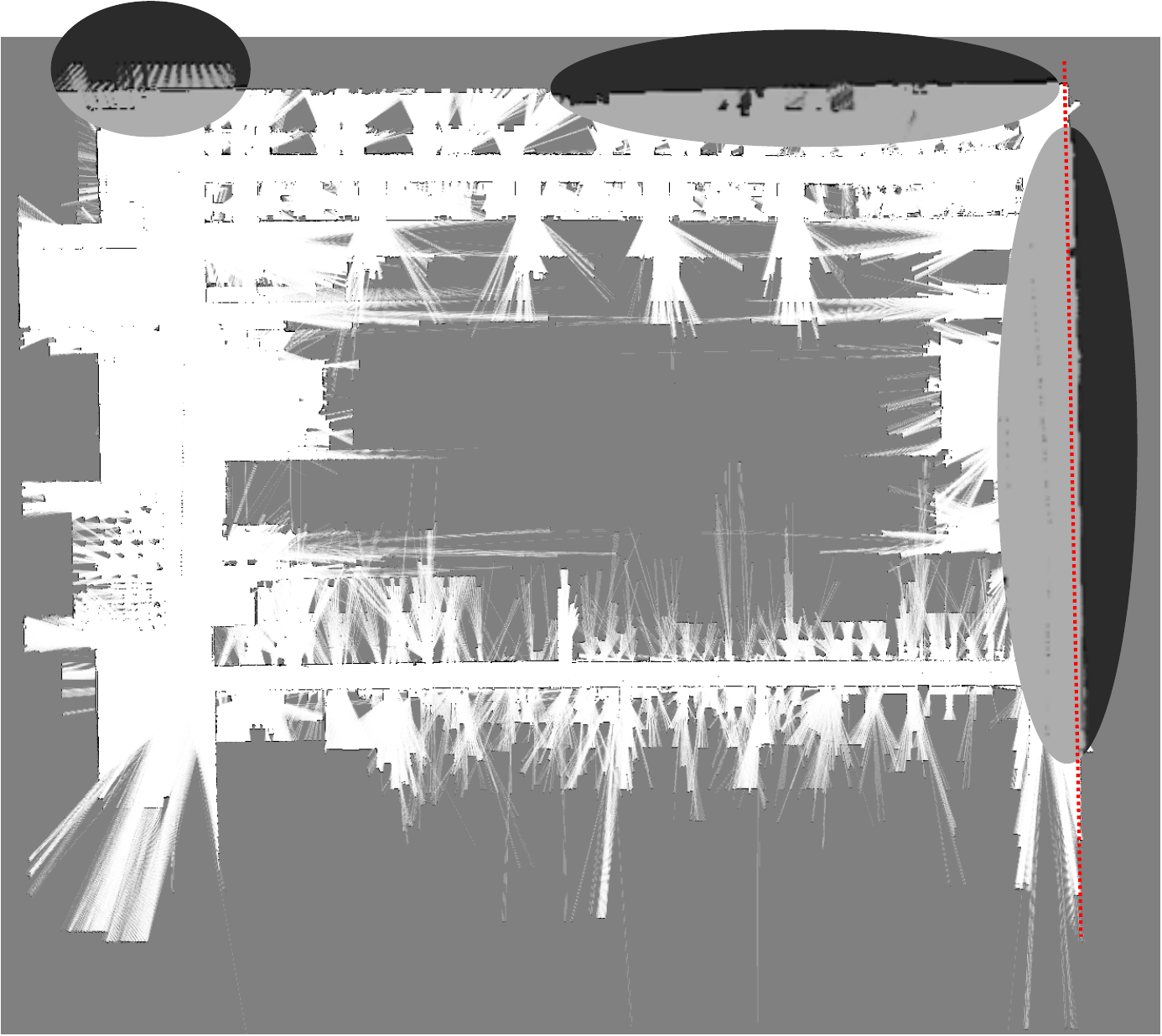}}

\centering \subfigure[b2 Our] {\label{fig_large_b2_Our}
\includegraphics[width=0.311\textwidth]{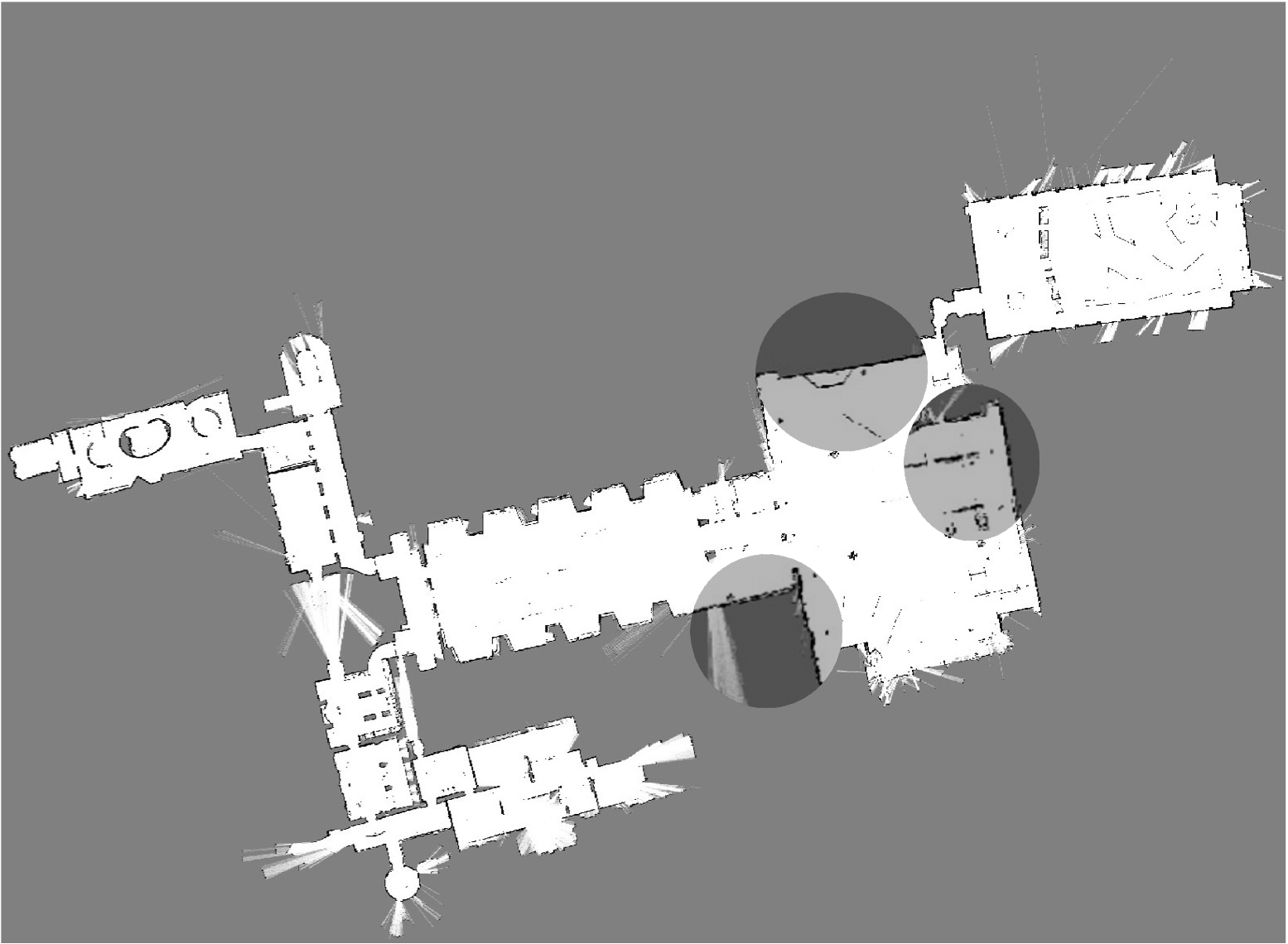}}
\centering \subfigure[b0 EG Our] {\label{fig_large_b0_Our}
\includegraphics[width=0.385\textwidth]{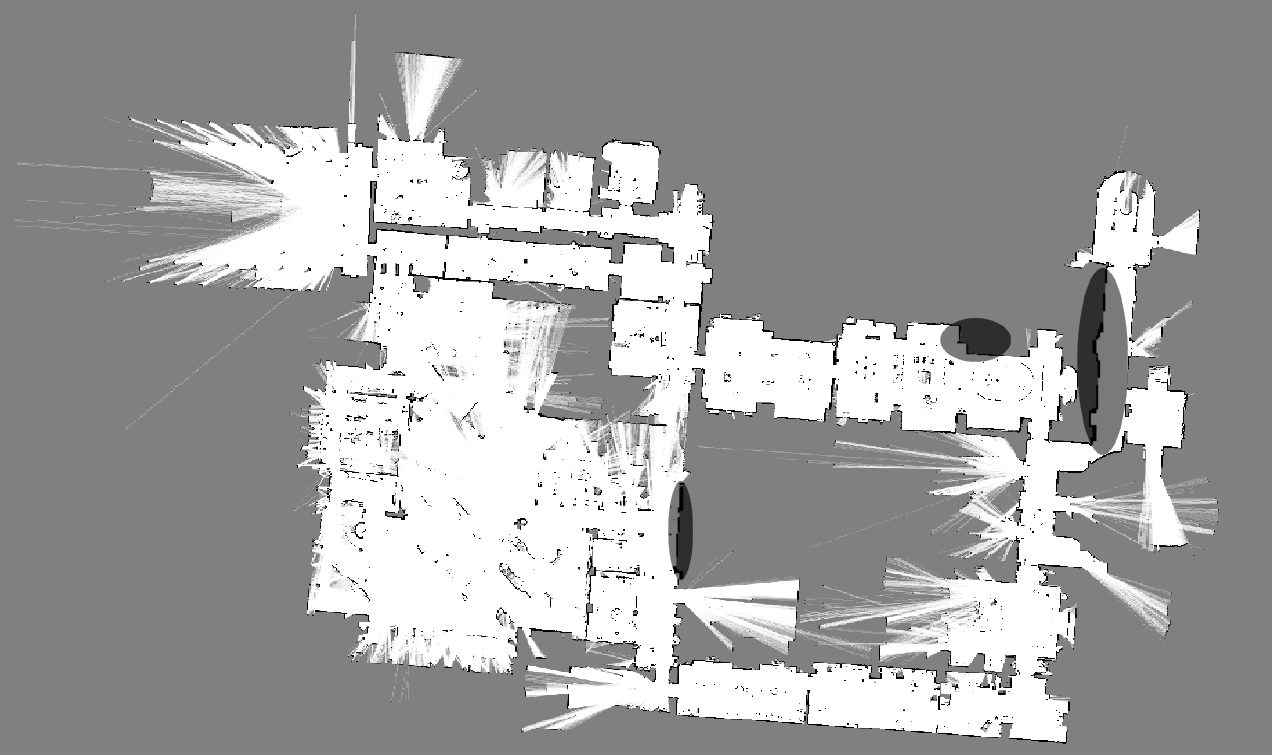}}
\centering \subfigure[C3 Our] {\label{fig_large_C3_Our}
\includegraphics[width=0.2625\textwidth]{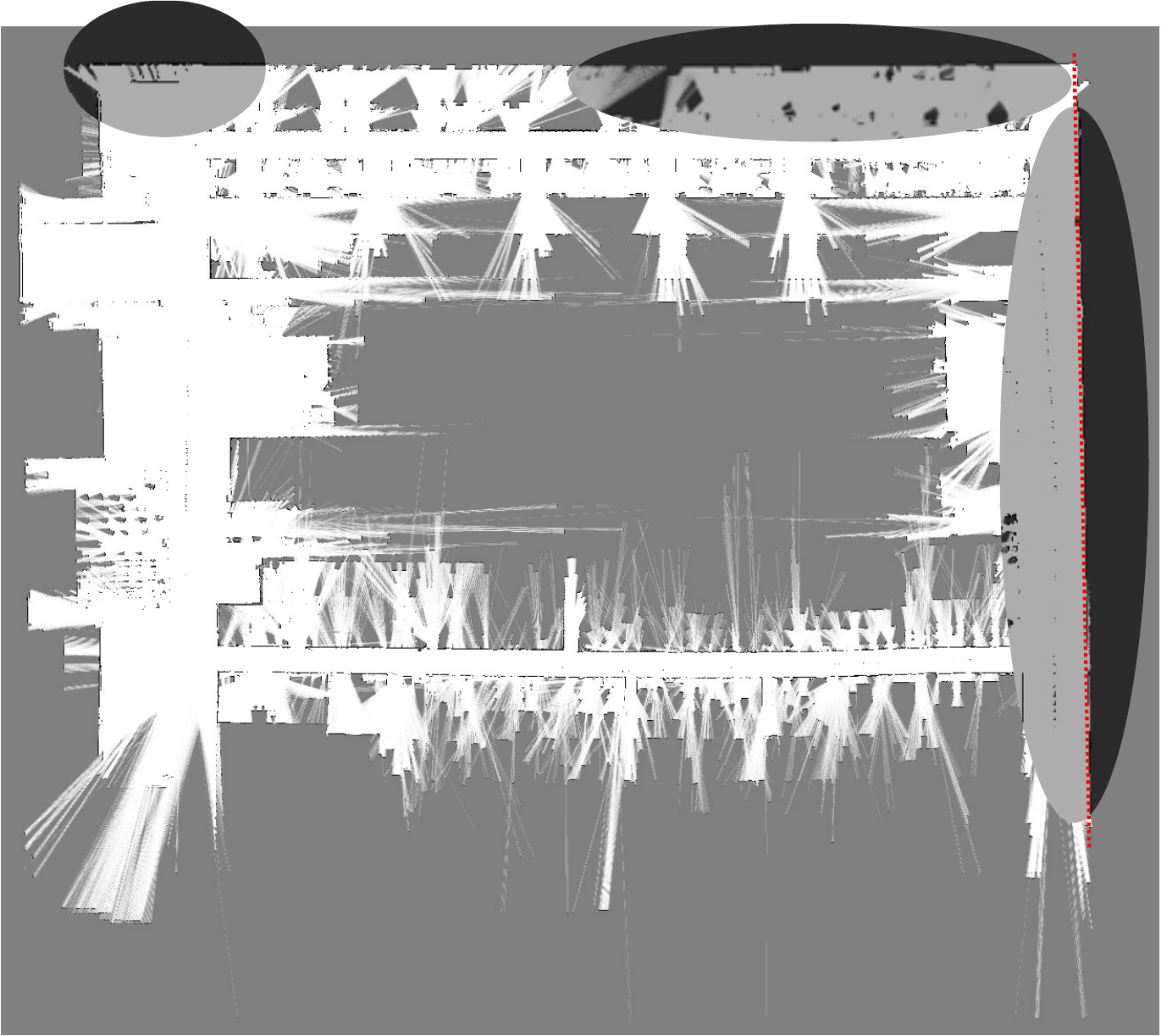}}
\caption{\label{fig_large_environment} \textit{Comparison of the results of our method and Cartographer in three large-scale environments.} The highlighted areas show our method outperforms Cartographer. The red dotted lines in (c) and (f) are reference lines, and it can be seen that the right wall of Cartographer's result is more curved while our result is closer to the straight line. Occupancy-SLAM fails with such large-scale environments.}
\end{figure*}

Our method can handle large-scale environments well, which is impossible with Occupancy-SLAM due to its increasing computational complexity and tendency to fall into local minima as the trajectory and map size grow. To validate the effectiveness of our algorithm in large-scale environments, three large-scale environment datasets are utilized, namely \textit{Deutsches Museum b0 EG}, \textit{Deutsches Museum b2}, and \textit{C3}, and their map sizes range from 150 m $\times$ 125 m to 250 m $\times$ 200 m. Here, we only compared our method with Cartographer, the results of OGMs are shown in Fig. \ref{fig_large_environment}, and it can be observed that our results are better than Cartographer, especially for Fig. \ref{fig_large_C3_Carto} and Fig. \ref{fig_large_C3_Our}, where the right wall of our results are clearly closer to the straight line whereas that of the Cartographer's results are somewhat curved.

\subsection{Time Consumption}
We use all the six practical datasets to evaluate the time consumption of our approach, Cartographer and Occupancy-SLAM, the results are presented in Table \ref{table_time_compare} (all elapsed times are evaluated using an Intel Core i7-1370P processor). The huge time consumption of Occupancy-SLAM is due to the high computational complexity of batch optimization using all the scans. Thus, it cannot handle the three large-scale datasets. In contrast, the proposed submap joining algorithm is very efficient, taking less than 1 second for the three normal-scale datasets and around 2 to 6 seconds for the three large-scale datasets. Since Cartographer and Occupancy-SLAM can solve the complete OGM based SLAM problems from scans, we also list the elapsed time of submap building. The total time of submap building and joining is still faster than that of Cartographer on most datasets. It should be mentioned that this total time consumption can be further reduced by using other efficient submap building methods, and the submaps can also be built out-of-core as in \cite{ni2007tectonic}.
\begin{table}[h]
		\centering
		\caption{Time Consumption (\textit{in seconds}) of Algorithms}
		\label{table_time_compare}
		\setlength{\tabcolsep}{1.2mm}{
		\begin{tabular}{lccccc}\toprule
        Dataset & Carto  &Occupancy-SLAM & \multicolumn{3}{c}{Ours} \\
			&  & & Submap & Joining & Total\\ \hline
		Car Park & 168  & 17138 &  133 &0.8 & 133.8\\
		Museum b0 1G& 152 & 26901 & 113 & 0.6 &113.6\\
		C5 & 146  & 27452 & 157 &0.6&157.6\\
        Museum b2& 1424 & - & 1248 &2.4 &1250.4 \\
        Museum b0 EG & 615 & - & 539 & 1.7 & 540.7\\
        C3 & 610 & - &737 & 5.2 & 742.2\\
  \hline
		\end{tabular}
		}
  \vspace{1.0em}
\end{table}

The high efficiency of the proposed map joining method is mainly due to the independent property of our formulation. Because of this, the pose-only GN method can be used to solve the NLLS problem efficiently. We compare our pose-only GN with the full GN where both poses and global map are solved at the same time, and the results are depicted in Table \ref{table_time_compare_GN}. It is clear that the calculation speed of our pose-only algorithm is much faster than the full GN algorithm.

\begin{table}[h]
		\centering
		\caption{Time Consumption (\textit{in seconds}) of Each Iteration*}
		\label{table_time_compare_GN}
		\setlength{\tabcolsep}{3.7mm}{
		\begin{tabular}{lccc}\toprule
		Dataset	& Ours & Full GN & Acceleration Rate \\
  \specialrule{0em}{2pt}{2pt}\hline
  \specialrule{0em}{0.5pt}{0.5pt}
		Car Park & 0.062 & 0.174 & 281\%\\
  \specialrule{0em}{0.5pt}{0.5pt}
		Museum b0 1G& 0.044 & 0.098 & 223\%\\
  \specialrule{0em}{0.5pt}{0.5pt}
		C5 & 0.036 & 0.058 & 161\%\\
  \specialrule{0em}{0.5pt}{0.5pt}
        Museum b0 EG& 0.157 & 0.419 & 267\%\\
        \specialrule{0em}{0.5pt}{0.5pt}
        Museum b2 & 0.226 & 0.378 & 167\%\\
        \specialrule{0em}{0.5pt}{0.5pt}
        C3 & 0.096 & 0.174 & 181\%\\
  \hline
    \specialrule{0em}{1pt}{1pt}
		\end{tabular}
  \begin{tablenotes}
     \item *Since our algorithm is equivalent to the full GN algorithm and thus has the same number of iterations, we only compare the time consumption per iteration.
   \end{tablenotes}
		}
\end{table}

\section{Conclusions}\label{sec:conclusion}

 This paper formulates the grid-based submap joining problem as a non-linear least squares problem to simultaneously optimize the global occupancy map and local submap frames, which is different from other submap joining approaches that only optimize poses by constructing a pose graph. More importantly, we show that for our formulation the existence of a special independent property helps solve this problem very efficiently by a pose-only Gauss-Newton algorithm. Our local submap joining approach is evaluated using datasets generated from two simulated environments and six practical datasets, demonstrating that it outperforms state-of-the-art algorithms in terms of speed and accuracy.

 This paper is the first to simultaneously optimize the global occupancy map and local map coordinate frames for the grid-based submap joining problem. In the experiments of this paper, we have used 3-10 local submaps depending on the scale of datasets. The best number of local maps for map joining (to achieve the best efficiency) and the best way to partition the whole laser dataset for building different high-quality local maps require further investigation, and these are left to our future work. We also plan to extend our approach to 3D cases to perform large-scale 3D grid map based SLAM more accurately and efficiently.

\bibliographystyle{IEEEtran}
\bibliography{bibliography}
\vfill

\end{document}